\PassOptionsToPackage{hyphens}{url}
\documentclass[msom,nonblindrev]{informs3_pom}

\OneAndAHalfSpacedXI

\usepackage{siunitx}
\usepackage{booktabs}
\usepackage{threeparttable}
\usepackage{float}
\usepackage{amssymb}            
\usepackage{amsfonts}
\usepackage{amsmath}
\usepackage{bm}
\usepackage{array}
\usepackage{multirow}
\usepackage{graphicx}
\usepackage{optidef}
\usepackage[export]{adjustbox}
\usepackage{graphicx}
\usepackage{siunitx}
\usepackage{csquotes}
\usepackage{makecell}
\usepackage{placeins}

\usepackage{natbib}
 \bibpunct[, ]{(}{)}{,}{a}{}{,}%
 \def\bibfont{\small}%

\usepackage{afterpage}
\usepackage{pdflscape}

\newcommand{\indep}{\perp \!\!\! \perp}

\usepackage{xspace}
	\newcommand\ie{i.\,e.\xspace}
	\newcommand\eg{e.\,g.\xspace}

	\usepackage{siunitx}
    \def\sym#1{\ifmmode^{#1}\else\(^{#1}\)\fi}
    \DeclareSIUnit\eur{\officialeuro}
    \DeclareSIUnit\M{M}
    \DeclareSIUnit\k{k}
	
	\usepackage[%
  sort&compress
]{cleveref}
  \crefname{chapter}{section}{sections}
	\Crefname{chapter}{Section}{Sections}
	\Crefname{figure}{Figure}{Figures}

\usepackage{comment}

\usepackage{chemmacros}
\usepackage{fontawesome}

\makeatletter
\renewcommand{\fps@figure}{H}         
\renewcommand{\fps@table}{H}         
\makeatother 

\usepackage{etoolbox}
\robustify\bfseries

\TheoremsNumberedThrough     

\EquationsNumberedThrough    

\begin{document}



\RUNAUTHOR{De-Arteaga, Feuerriegel, and Saar-Tsechansky}

\TITLE{Algorithmic Fairness in Business Analytics:\\ Directions for Research and Practice}

\RUNTITLE{Algorithmic Fairness in Business Analytics}

\ARTICLEAUTHORS{%
\AUTHOR{Maria De-Arteaga}
\AFF{The McCombs School of Business, The University of Texas at Austin, 2110 Speedway, Austin, Texas 78705, USA}
\EMAIL{dearteaga@utexas.edu}
\AUTHOR{Stefan Feuerriegel}
\AFF{LMU Munich, Geschwister-Scholl-Platz 1, 80539 Munich, Germany}
\EMAIL{feuerriegel@lmu.de}
\AUTHOR{Maytal Saar-Tsechansky}
\AFF{The McCombs School of Business, The University of Texas at Austin, 2110 Speedway, Austin, Texas 78705, USA}
\EMAIL{maytal@mail.utexas.edu}
} 

\ABSTRACT{%

\vspace{-.5cm}

\noindent
{\OneAndAHalfSpacedXI
The extensive adoption of business analytics (BA) has brought financial gains and increased efficiencies. However, these advances have simultaneously drawn attention to rising legal and ethical challenges when BA inform decisions with fairness implications. As a response to these concerns, the emerging study of \emph{algorithmic fairness} deals with algorithmic outputs that may result in disparate outcomes or other forms of injustices for subgroups of the population, especially those who have been historically marginalized. Fairness is relevant on the basis of legal compliance, social responsibility, and utility; if not adequately and systematically addressed, unfair BA systems may lead to societal harms and may also threaten an organization’s own survival, its competitiveness, and overall performance. This paper offers a forward-looking, BA-focused review of algorithmic fairness. We first review the state-of-the-art research on sources and measures of bias, as well as bias mitigation algorithms. We then provide a detailed discussion of the utility-fairness relationship, emphasizing that the frequent assumption of a trade-off between these two constructs is often mistaken or short-sighted. Finally, we chart a path forward by identifying opportunities for business scholars to address impactful, open challenges that are key to the effective and responsible deployment of BA.}

}

\HISTORY{}

\maketitle

\sloppy
\raggedbottom

\vspace{-1cm}

\section{Introduction}
\label{sec:intro}

The widespread adoption of business analytics (BA) has brought financial gains and increased efficiencies \citep{Choi.2018,Cohen.2018,Cui.2018}. However, cultural and corporate transformations have led to increased focus on other important consequences of analytic-based decision-making \citep{Nkonde.2019}. In particular, the once dominant view that a company's responsibility is principally to produce profit for shareholders has meaningfully shifted, culminating in the 2019 Business Roundtable\footnote{\SingleSpacedXI\scriptsize The Business Roundtable is the leading business lobby in the United States, with CEOs representing approximately 30\,\% of the U.\,S. market capitalization. URL: \url{https://www.businessroundtable.org/business-roundtable-redefines-the-purpose-of-a-corporation-to-promote-an-economy-that-serves-all-americans}} statement that corporations share a key responsibility to \emph{all stakeholders}: customers, employees, suppliers, and communities--in addition to shareholders. The statement conveys a vision where, in order to deliver value to stakeholders, firms' decisions and actions must also reflect broadly-held societal values, including in particular fairness, explicitly noted in the statement. This important transition in corporate goals underscores a realization that aligning firms with widely-held notions of ethics and fairness is crucial for the long-term well-being of firms and of the economy. Globally, improving fairness and reducing inequality are viewed as integral to sustainable development, as also reflected by the United Nations' Sustainable Development Goals \citep{SDG.2015}.

BA is assuming a major role in optimizing company operations and decision-making \citep{Choi.2018,Cohen.2018,Cui.2018}; yet, the BA literature has thus far focused largely on traditional business utilities \citep{Cachon.2020}. The transition towards incorporating notions of fairness into BA is necessary but non-trivial: it requires understandings and appreciation of an array of complex challenges, as well as concentrated efforts to advance the state-of-the-art so as to achieve meaningful impact in practice. Fairness in algorithmic decision-making refers to models' outcomes systematically deviating from statistical, moral, or regulatory standards \citep{Danks.2017}, especially in ways that disproportionately impact groups along social axes, such as race, gender, and class \citep{Mitchell.2021}. Decisions informed by BA intended to benefit traditional business utilities can simultaneously undermine fairness, ultimately introducing ethical, regulatory, and business risks, including to a firm's competitiveness, and overall performance. 

Evidence of how BA can undermine fairness and lead to discrimination has been reported across different areas. For example, to inform revenue management, the American retailer Staples adopted a data-driven online pricing algorithm, that adjusted prices based on customers' proximity to competitors' stores \citep{WSJ.2012}. This data-driven pricing, however, led to higher prices for low-income customers, as these customers often lived farther from competitors' stores. In operations management (OM), allocations of resources may be not be equal among customers, suppliers, or other stakeholders, thus making fairness a common concern \citep{Rea.2021}. In healthcare operations, state-of-the-art algorithms for risk scoring were found to systematically underestimate the health risk of black patients as compared to equally sick white patients \citep{Obermeyer.2019}. As we will discuss at length below, the ultimate goal of research towards methods for both detection and mitigation of such biases in BA is to yield consistently fair business operations. 

To date, BA practices in selected domains have been regulated in the United States, as well as in other countries \citep[cf.][]{Barocas.2016,Kleinberg.2018}. In the U.\,S., for example, fair lending laws penalize some forms of algorithmic discrimination in risk scoring. Similarly, the Equal Credit Opportunity Act prohibits certain types of algorithmic discrimination by age or race, and the Algorithmic Accountability Act requires firms to conduct assessments of high-risk systems involving automated decisions, including specifically systems that may contribute to bias or discrimination. In the European Union, accountability of analytics is directed by the General Data Protection Regulation~(GDPR) and, in the future, will be further subject to an Artificial Intelligence Act. Overall, there is a growing tendency by regulators around the world to enact laws that forbid disparate treatments, including specifically in the context of algorithmic decision-making \citep{WhiteCase.2020}. 

Regulatory and societal pressures over the risks of algorithmic bias and discrimination have led to an emerging body of work on \textbf{\emph{algorithmic fairness}}. These works focus on understanding the different sources of bias, and on effective data-driven techniques for both detecting and mitigating algorithmic bias \citep[\eg,][]{Chouldechova.2017,CorbettDavies.2017,Dwork.2012,Hardt.2016}. For research and practice, algorithmic fairness research is key to identify sources of potential unfairness in important contexts in which BA informs decisions and underlies operations, and to develop and evaluate the implications of methods to detect and mitigate bias in these contexts. The potential of BA to improve business operations and decision-making offers ample opportunities to impact the transition into a business culture that aims to deliver value to all stakeholders. We hope that this paper will contribute to this transition by facilitating understandings of the challenges ahead, and that it will inspire future BA research that will advance algorithmic fairness in business operations.

Most algorithmic fairness research thus far has been driven by a computer science research agenda \citep[\eg,][]{Cohen.2019,Dwork.2012,Friedler.2016,Kleinberg.2017,Pleiss.2017}, and existing surveys provide perspectives from computer science \citep{CorbettDavies.2018, Chouldechova.2020}, statistics \citep{Mitchell.2021}, philosophy \citep{Danks.2017,Fazelpour.2021}, law \citep{Barocas.2016}, and economics \citep{Cowgill.2020}. Tutorials have also provided summaries of the literature on sources of bias and methodologies to detect and mitigate it \citep{Fu.2020}. 

Different from existing surveys, our paper is the first to focus on algorithmic fairness in the context of BA/OM, and, to this end, we discuss several opportunities and challenges that are unique to business operations. In particular, we discuss how bias may arise in key areas such as revenue management, retail operations, logistics, and supply chain management; we provide a structured and novel characterization of the utility-fairness relationship that it fundamental to business operations; and we identify key directions for impactful future work that build on the strengths of the BA/OM communities.

\begin{figure}
\centering
\vspace{-0.3cm}
\includegraphics[width=0.6\linewidth]{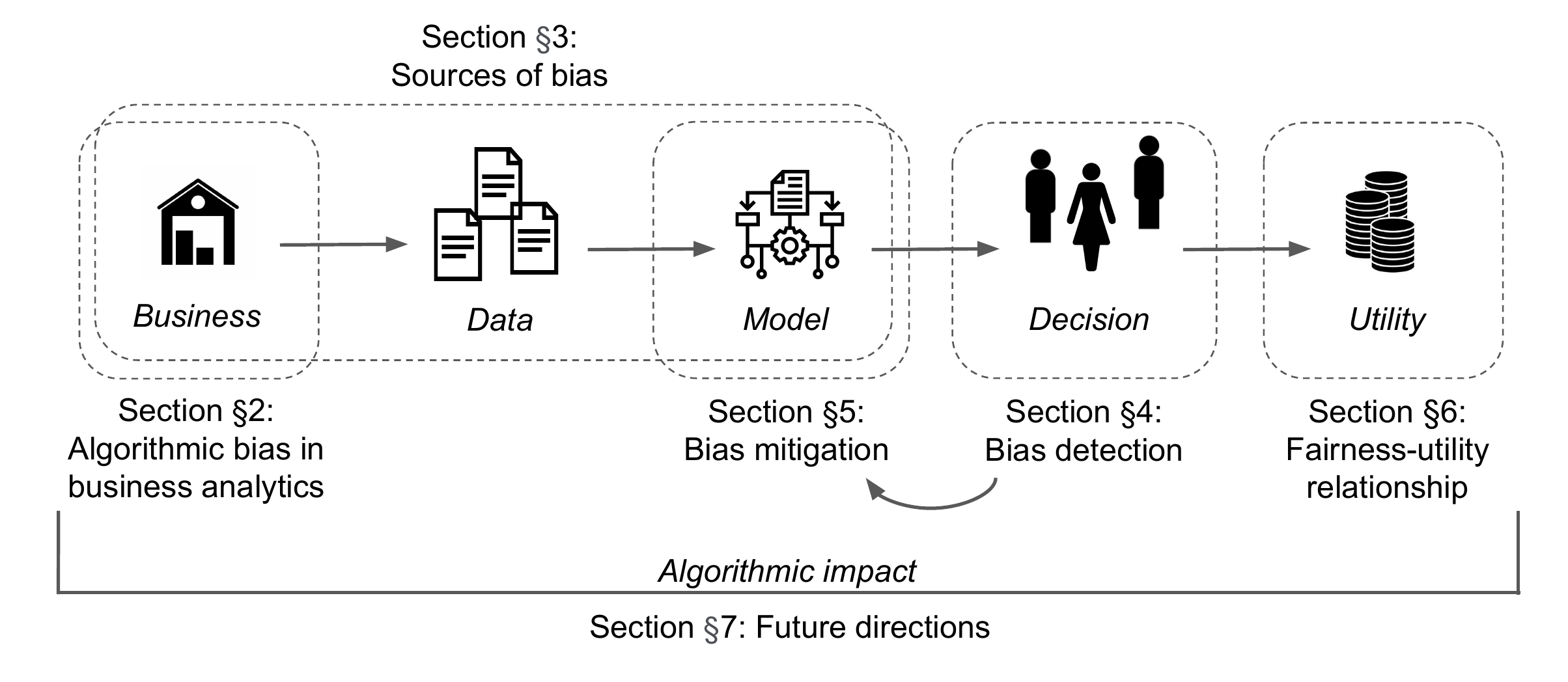}
\vspace{-0.3cm}
\caption{Overview of the paper outlining how algorithmic fairness can be addressed in business analytics.}
\label{fig:overview}
\vspace{-0.6cm}
\end{figure}

In this paper, we first review evidence of algorithmic bias in business operations in \Cref{sec:bias_examples}, followed by a taxonomy of the sources from which biases may arise in~\Cref{sec:bias_sources}. We then review the state-of-the-art in both algorithmic bias detection (\Cref{sec:bias_detection}) and algorithmic bias mitigation (\Cref{sec:bias_mitigation}). In \Cref{sec:fairness-utility}, we discuss the critical relationship between bias and traditional business objectives (\eg, profits), characterizing where there may or may not be a trade-off. Finally, we outline important future directions for BA (\Cref{sec:future_directions}), critical to ensuring the impact of BA in practice. \Cref{fig:overview} offers an overview of the structure of the paper.

\section{Algorithmic Bias in Business Analytics}
\label{sec:bias_examples}

In this section, we discuss fairness issues that arise in specific BA tasks. Specifically, we review examples that underscore the diverse contexts in operations and management where algorithmic decisions can be inflicted with bias, the variety of reasons that can lead to algorithmic bias, and the different kinds of individual and societal implications of such biased treatments (see \Cref{tbl:examples} for further examples).

In service operations, the use of BA to assess creditworthiness has been found to present algorithmic bias risks \citep{Hardt.2016}. Historical data is often used to model and infer creditworthiness; yet, such data might already reflect human or societal biases. Consequently, unless such biases can be detected and mitigated, they may be encoded into models that inform future decisions \citep{Kozodoi.2021}. For instance, it has been shown that historical decisions on creditworthiness can be biased against people of certain age, race, and gender, even if the sensitive attributes are unobserved by the model \citep[][]{Kallus.2021}. In part, this is because other attributes may act as proxies for sensitive attributes, such as gender; for instance, salary may be a proxy for gender when men receive a higher salary than women \citep{Pierce.2021}. From an ethical perspective, such disparities in the access to credit are especially harmful, given that credit (\eg, student loans) is an essential instrument for enabling socioeconomic opportunities and economic mobility.

A meaningfully different case of algorithmic bias in the context of logistics has risen when Amazon relied on BA to determine which U.\,S. regions are offered free same-day delivery \citep{Bloomberg.2016}. These data-driven decisions, which prioritized zip codes with high concentration of Amazon Prime members, ultimately limited access to the service for individuals in predominately black neighborhoods, even when all surrounding neighborhoods were offered the service. The resulting disparities compounded existing barriers for residents of predominantly black neighborhoods, who are already disproportionately impacted by food deserts \citep{Bower.2014}. As Amazon's role as a main provider of basic goods (\eg, fresh food) grows, the societal value of providing equal access to delivery services becomes more salient.  

In healthcare operations, analytics is increasingly leveraged to stratify patients by risk, which subsequently informs healthcare resource allocation \citep[\eg,][]{Ayvaci.2017,Ayvaci.2018,Helm.2011,Ibrahim.2016}. For instance, analytics is used to estimate patients' risks of mortality, adverse reactions to a treatment, or hospital readmission. Such risk scores are used to inform treatment decisions so as to achieve cost-effective allocation of resources. Particularly with the growing use of machine learning \citep{Bastani.2022}, healthcare analytics has been highly effective in enabling efficiencies in the U.\,S. and other countries, and have also been shown to reduce disparities in some contexts \citep{Ganju.2020}. However, state-of-the-art algorithms were also found to exhibit racial bias, systematically inferring that black patients were less sick than they actually were \citep{Obermeyer.2019}. In this case, the bias stemmed from the use of health costs as a proxy for health needs, without considering that less money is spent on black patients who have the same level of need.

\begin{table}[tbp]
\centering
\caption{Examples where algorithmic bias occurs in business analytics. }
\label{tbl:examples}
\SingleSpacedXI
\tiny
\def\arraystretch{1.3}
\begin{tabular}{p{2.5cm} p{13.5cm}}
\toprule 
\textbf{Area} & \textbf{Example} \\ 
\midrule 
Revenue management & Dynamic pricing identifies a user's propensity of purchase based on available covariates. However, when the geographic location of users is considered, this might bias the outcome against certain neighborhoods and, as a consequence, against high/low income households.\\
Service operations & Credit scoring is used to determine when customers can use credit vs. debit cards or whether they are offered a loan. However, some of these algorithms have been found to be implicitly biased against women and racial minorities \citep{Hardt.2016}. \\
Retail operations & Allocating marketing efforts to customers is based on algorithmic tools. However, Google Ads was found to be more likely to show high-paying job ads to men than to women \citep{Datta.2015}.\\
Logistics & Amazon used business analytics to determine where it should offer free same-day delivery. However, the data-driven decisions resulted in the exclusion of primarily black neighborhoods.\\
Supply chain management & Analytics supports the distribution of COVID-19 vaccines to those who need them most. However, simply targeting all people alike is biased toward high-income people, since low-income people are often at higher infection risk. Hence, analytics must correct for that to target the most vulnerable populations effectively \citep{IBM.2021}.\\ 
Human resource operations & Business analytics is increasingly used in hiring to screen job applications and then make decisions for workforce selection. Amazon's algorithmic tools for automating hiring decisions systematically favored men applicants over women \citep{Dastin.2018}. \\
Healthcare operations & Analytics for treatment planning in diabetes management is often calibrated toward white, middle-class patients. Because of this, analytics may introduce over-proportional high error rates for other patient cohorts \citep{Barda.2021}.\\
Public sector operations & Decision models for policing rely commonly on data from past arrests, thereby replicating existing racial biases in policing operations \citep{Lum.2016}. \\ 
\bottomrule
\end{tabular}
\end{table}

In addition to the real-world examples uncovered by researchers and reporters, theoretical research in OM has also studied fairness considerations in optimization, especially with regard to the ``price of fairness'' \citep{Bertsimas.2011,Bertsimas.2012, Bertsimas.2013,Cohen.2019,Katok.2014,Rea.2021}. For example, the trade-offs brought about by fairness have been studied for single-resource allocations \citep{Bertsimas.2011} and for multi-resource allocations \citep{Rea.2021}. In a related context, \citet{Katok.2014} analyze how supply chain coordination is affected by fairness in wholesale pricing. In healthcare operations, prior literature has used theoretical models to understand the impact of different fairness constraints on organ transplantation \citep{Bertsimas.2013}, appointment scheduling \citep{Qi.2017}, and pricing \citep{Cohen.2019}. 

In spite of this growing body of work and evidence, algorithmic decision-making often enjoys a veneer of inherent objectivity, deemed an evidence-based alternative to biased and idiosyncratic human decisions \citep[\eg,][]{Cui.2020,Pierce.2021}.\footnote{\SingleSpacedXI\scriptsize A growing stream of research in OM contributes evidence of bias in human decision-making. For instance, in bargaining, women consistently receive outcomes below their productivity level, whereas men are consistently overcompensated \citep{Pierce.2021}. In the context of the sharing economy, \citet{Cui.2020} conducted randomized field experiments with 1,801 hosts at Airbnb. Requests from guests with African-American sounding names are 19.2 percentage points less likely to be accepted than requests from people with white-sounding names, thus finding evidence of discrimination.} The above examples suggest that algorithmic bias is widespread in the deployment of BA and that it can arise across a variety of contexts and for different reasons. Moreover, the risk of algorithmic bias will increase with a more prevalent use of big data among business and operations. This presents ample opportunities for BA scholars and practitioners to meaningfully contribute to algorithmic fairness, and establish best practices for fair conduct towards suppliers, customers, employees, and society at large.

\FloatBarrier
\section{Sources of Bias in Business Analytics}
\label{sec:bias_sources}

To understand the risks and consequences of algorithmic bias, one must first understand where such bias may come from. Fairness concerns the allocation of goods and burdens, and thus is a property of the decisions that may be made or informed by algorithms. As such, bias in algorithmic-informed decisions may stem from biased algorithmic outputs, or from other portions of the sociotechnical system that the algorithm is embedded on \citep{Dolata.2022}. In this section, we provide a taxonomy of common sources of algorithmic bias--understood as biased algorithmic outputs--with an emphasis on business operations problems. We then conclude with a discussion of other sources of bias during deployment. Throughout, we consider algorithms that estimate a model that maps input onto a target outcome. 

\subsection{Data Collection and Representation}
\label{sec:sources_data}

\subsubsection{Sampling bias.}
\label{sec:sampling_bias}

While the possibility of learning from low-cost, observational, and easily available data is often listed as an advantage of data-driven decision-making when pitching BA solutions, the importance of \emph{good data}, and not just \emph{big data}, should not be undervalued. Estimation and evaluation of decision models rely on assumptions about the data distribution being representative of a distribution of interest. When there is a mismatch between the data distribution upon estimating a decision model and its deployment (\ie, a domain shift), this can undermine the generalizability of the model. Sampling bias has important fairness implications, as convenient data samples do not represent all subpopulations equally. In particular, marginalized groups who have been historically excluded or underserved are frequently under-represented in data. For instance, clinical trials for learning personalized treatment rules involve primarily white patients \citep[cf.][]{Warren.2020}; therefore, the estimated treatment rule might be ineffective when deployed in a population that also comprises patients of other races. 

Under-representation in data is a pervasive phenomenon across many domains. For instance, marketing operations are confronted with a sampling bias as targeting is often calibrated using data of a subset of the user base (\eg, a specific geographic area or a carrier route) which might not be representative of the entire user base \citep{Simester.2020}. When exploring the use of alternatives sources of data that can provide business value such as the use of social media data to improve sales forecasts \citep{Cui.2018}, it is essential to consider the potential limitations stemming from sampling bias. For instance, while the availability of social media offers the possibility to listen to what people are saying, it is crucial to be cautious about the biases in terms of whose views and voices are included in such data \citep{Malik.2015}.

Finally, an important note ought to be made regarding the link between data sampling and algorithmic bias. A representative sample, collected by randomly sampling from the population of interest, may also lead to algorithmic bias. Algorithms often optimize for overall performance, which means that generalization from such a sample may lead the algorithm to prioritize the majority over the minority populations. We discuss this in further detail in \Cref{sec:sources_objective}. 

\subsubsection{Differential subgroup validity.}
\label{sec:differential_subgroup_validity}

Sampling bias refers to \emph{who} or \emph{which} instances are represented in the data; \emph{how} they are represented also matters. In a modeling task, the predictive or prescriptive power of covariates relative to an outcome of interest may differ across groups \citep{CorbettDavies.2018}, which may be a source of disparate performance.

The choice of which covariates to collect or use may be informed by practitioners' intuition and domain expertise, as well as previous studies that have established associations between certain covariates and the outcome of interest. As a result, the choice of covariates may be overly influenced by relationships that exist for advantaged groups. Meanwhile, covariates that are indicative of an outcome of interest for minority or historically marginalized groups may be missed. For instance, consider the growing industry of artificial intelligence for health management aimed at allocating preventive care to patients by estimating the likelihood of a condition. Here, the strength of the correlations between covariates and outcomes may vary across groups. For example, dermatological symptoms may present differently across races; asking patients whether they have a \textquote{red rash} corresponds to collecting and relying on a covariate that only has predictive power for certain skin colors \citep{Adamson.2018,Ebede.2006}, and could result in a model that has a subpar performance for patients with dark skin.

Moreover, even covariates that initially have the same predictive power may become differentially predictive if only some groups can strategically adapt to the incentives introduced by the utilization of a decision model. Consider the use of standardized tests in college admission decisions. The use of tests as a criteria for admissions motivates students to invest in tutoring and to retake tests. However, not all students can afford this. This strategic adaptation to incentives, sometimes referred to as \textquote{strategic manipulation,} may lead to variations in predictive and prescriptive power across groups and lead to algorithmic bias \citep{Hu.2019}.

While the choice of covariates with differential predictive power is often the result of (harmful) oversight, it has been noted in the literature that this could also be intentionally used as a mechanism for \textquote{algorithmic redlining.} For example, unfairly limiting access to credit to minority applicants could be achieved by basing estimates only on coarse information, like one’s neighborhood, and ignoring individual-level factors, like credit history \citep{CorbettDavies.2018}.

\subsubsection{Biased observed outcomes.} 
\label{sec:biased_observed_outcomes}

Finally, \emph{what} an algorithm predicts may also be a source of bias. At this stage, we differentiate between bias that stems from a gap between the outcome of interest and the measured outcome, and bias that stems from relying on human assessments as target outcomes.  

\paragraph{Proxies and mismeasurement errors.}

Decision-makers pursue complex, multidimensional, and, at times, not easily quantifiable objectives. Furthermore, getting the most value from data often means repurposing data that was collected for other reasons, in which case the quantified and recorded information may not be perfectly suitable for what the new task requires. Consequently, a common step in BA is to choose proxies that can approximate and take the place of constructs of interest \citep{Bastani.2021c}. For example, the goal of hiring the most qualified candidates may be translated into the task of hiring candidates based on their predicted future performance reviews or predicted future sales \citep{Raghavan.2020}, and the goal of prioritizing high-risk patients for early medical care may be transformed into the task of estimating which patients are most likely to incur high costs \citep{Obermeyer.2019}. 

Proxies, as their name indicates, are not a perfect match with the desired objective. Importantly, the extent to which they diverge from the objective--and the way in which they do--may vary for different subpopulations, which can be a source of algorithmic bias. A prominent example was studied by \citet{Obermeyer.2019}, who analyzed a healthcare algorithm deployed nationwide in the U.\,S. as part of a high-risk care management program. The model's goal was to estimate patients' health needs and inform which patients should be prioritized for enrollment in the program. Rather than directly estimating \emph{needs}, the model predicted health \emph{costs}. Intuitively, the health costs a patient incurs would be reflective of health needs and diseases, making cost a seemingly reasonable and easily observable proxy. However, unequal access to care means that less money is spent on black patients despite having the same level of needs as other patients; thus, the algorithm underestimated the needs of black patients, and erroneously concluded that they were healthier than equally sick white patients. 

In this example, the use of cost as a proxy was partly motivated by the fact that \textquote{health needs} is a complex and ambiguously defined concept, thereby difficult to meaningfully quantify as a predictive goal. But even when the outcome of interest is \emph{well defined}, it may not be \emph{well measured}. For instance, consider BA used in the public sector to help inform the allocation of resources. In particular, work focused on police operations \citep[\eg,][]{Perry.2013} has proposed optimizing the allocation of police patrols by predicting which locations are likely to be so-called crime hotspots. Whether something is a crime has, in theory, a very clear legal definition. Yet, we do not always observe whether someone committed a crime; instead, we observe arrests or victim reports. Researchers have showed that relying on policing data to induce models perpetuates racial biases in policing practices and leads to feedback loops in police allocation \citep{Ensign.2018,Lum.2016}. Similarly, using victim reporting data to fit the models may lead to misallocations of policing resources both in the form of over-policing and under-policing due to differential crime reporting rates across areas \citep{Akpinar.2021}.

\paragraph{Human assessments.}

In many OM domains, human assessments are a fundamental component of the data used in BA \citep{Geva.2021}. In particular, in the context of predictive modeling, it is often the case that the outcome variables used for estimating a model correspond to human assessments. Three types of human assessments are common: (1)~historical expert assessments available in organizational information systems, (2)~elicited expert assessments collected for the purpose of estimating a model, and (3)~crowdsourced labels. Whenever labels are provided by humans, there is a direct path through which societal biases can be encoded in a model.  

Relying on historical expert assessments can perpetuate past biases in an organization. For example, in human resource operations, developing a model to predict who is likely to be hired for a job, or who is likely to receive positive managerial reviews, could replicate past prejudices in hiring and promotion \citep{Dastin.2018}. This problem extends to domains in which we would like to think that there exists a \textquote{gold standard} or easily acquired \textquote{ground truth.} For instance, the potential use of automated radiological screening has recently made waves in the healthcare industry \citep{Choy.2018,McKinney.2020}. However, artificial intelligence used in radiology rely on experts' assessments, who routinely disagree on their interpretations, a problem referred to by experts as the lack of a histopathological gold standard \citep{Adamson.2019}. From human resource operations to healthcare devices, there are many business domains in which trained models rely on experts' assessments. Recognizing this, and understanding the potential biases encoded via this pathway, is a first step towards developing analytics that mitigate human biases \citep{Ahsen.2019}, instead of inadvertently perpetuating them under a veil of \textquote{objective} data-driven decisions.

Finally, with a growing industry of crowdwork and digital labor markets \citep{Allon.2020,Gray.2019}, BA practitioners have turned their attention to the potential value of labels provided by crowdworkers. Research has shown that assessments collected via crowdsourcing platforms are sensitive to societal biases \citep{Dressel.2018}. Whenever models are optimizing for \textquote{wisdom of the crowd,} the biases of the crowd can also be incorporated into the model.

\subsection{Model Estimation}
\label{sec:model_estimation}

\subsubsection{Optimization objectives and evaluation bias}
\label{sec:sources_objective}

Once the data is collected, the next stage of a BA task is to estimate a model. Here, bias may arise from optimization objectives that are too coarse and/or focused on narrowly defined criteria. Most commonly, the standard BA practice is to optimize for overall or aggregate performance over the population may lead the model to disregard minority groups, even in the absence of sampling bias. In the context of predictive modeling, efforts to avoid overfitting often take the form of using constrained classes of classifiers or introducing loss functions that penalize complexity through regularization. Hence, if different subpopulations are represented in different proportions, the observations belonging to the majority population will be more influential in \emph{what} functional representation is estimated, and the model might not generalize well to the minority population. Consider again the example of presentation of dermatological symptoms across different skin colors \citep{Adamson.2018}. The correlations that are predictive of a target of interest may vary across groups; thus, a model that optimizes for overall performance with a complexity constraint might only learn which are predictive symptoms for people with light skin color, if this group makes up for a larger proportion of the data. Thus, even if different subgroups are represented in the data in proportions that are consistent with the distribution that will be encountered upon deployment, optimizing for overall performance can result in algorithmic bias.

\subsubsection{Bias ingrained in the assumptions of the objective.} 
\label{sec:bias_assumptions_objective}

Bias may sometimes be inherently ingrained in implicit assumptions. For instance, businesses are increasingly turning to the use of computer vision and facial analysis as part of hiring processes, customer satisfaction, and, most recently, COVID-19 safety protocols. Racial and gender bias in these technologies have received widespread attention \citep{Buolamwini.2018}. However, services such as gender recognition, increasingly used in business operations as part of marketing and customer-centered efforts, make the assumption that a person's gender can be determined from their facial features, which may be inherently exclusionary of transgender individuals \citep{Hamidi.2018}. In recent years, there has also been a surge in the development and commercialization of technologies that attempt to use facial images to predict different personality traits \citep{McFarland.2019}, including some traits that are potentially relevant for hiring and human resource operations. The practice of inferring a person's character from their appearance is known as physiognomy, a debunked theory that was used as a basis for scientific racism \citep{Gould.1996} and has seen a revival through facial analysis with computer vision \citep{Bowyer.2020,Chinoy.2019}. In cases in which the task itself makes prejudiced assumptions, the goal of mitigating bias is an ill-posed task. Before setting out on the goal of mitigating algorithmic bias, it is fundamental to interrogate the implicit assumptions of the task definition and identify whether any of these may be \emph{inherently} biased or unethical.

\subsection{Deployment}
\label{sec:deployment}

\subsubsection{Bias in BA adoption.}

In some tasks, such as targeted advertisement, BA is used to make autonomous decisions. In many other settings, such as hiring, pricing and inventory management, human decision-makers may retain discretionary power. When and how experts make use of the information provided by a BA tool, and whether discretionary power improves or harms the quality and fairness of decisions, is conditioned on many factors. In the context of revenue management, empirical research has found that when data-driven models are used to assist pricing and merchants' decisions, discretionary power may be effective only for some types of products \citep{Kesavan.2020}, and when pricing for quotes or clients with unique or complex characteristics \citep{KarlinskyShichor.2019}. Similarly, in the context of ordering behavior in retail stores, empirical evidence suggests that store managers can improve upon an automated replenishment system by successfully integrating information that is unavailable to the algorithm \citep{vanDonselaar.2010}. Broadly speaking, decision-makers' likelihood to adhere to algorithmic recommendations relates to phenomena of algorithm aversion \citep{Dietvorst.2018} and automation bias \citep{Skitka.2000}. Importantly, the probability that a decision-maker will accept an algorithmic recommendation is not uniformly distributed across all observations. In particular, heterogeneity in adherence to recommendations across sensitive groups can be a source of increased disparities \citep[\eg,][]{Zhang.2021}. In the criminal justice system, differential adherence has been shown to exacerbate disparities in incarceration across black and white defendants \citep{Albright.2019}, and across socioeconomic brackets \citep{Skeem.2020}. Thus, even if an algorithmic output does not exhibit bias against a subgroup, variation in humans' likelihood to update their prior beliefs or to override the recommendations may result in disparate impact of algorithmic-informed decisions. Conversely, it is also possible that discretionary power of a human-in-the-loop approach could mitigate bias in BA. 

In this section, we presented a taxonomy of sources of bias that aims to provide readers with a lens through which to analyze potential biases in a given OM task. It is useful to note that, in practice, the sources of bias for a given dataset or task need not be unique, and instead multiple sources can be present simultaneously.

\section{Detecting Algorithmic Bias in Business Analytics}
\label{sec:bias_detection}

We now discuss how algorithmic bias can be detected in BA and how fairness is formalized mathematically. This yields so-called algorithmic notions of fairness, which can be used by researchers and practitioners to measure disparities in decision-making.

\subsection{Mathematical Formalizations of Algorithmic Fairness}

By formalizing fairness mathematically, it is possible to quantify the deviation from an outcome that would be considered fair, and then optimize for solutions that better meet this goal. However, fairness is an inherently socio-legal concept, with no universal agreed-upon definition. Interpretations of fairness may also often be context-specific \citep{Kleinberg.2018} and subject to disagreement. Moreover, there is no direct mapping of socio-legal notions of fairness onto mathematical formulations. While we review below several common notions of fairness from the literature, it is necessary to carefully consider the requirements of each BA tasks and, where needed, make corresponding adaptations. Because of this, an out-of-the-box approach to algorithmic fairness based on which one can audit BA and call it \textquote{fair} is precluded.

Algorithmic notions of fairness can be loosely grouped into three categories: (1)~notions that define fairness across groups; (2)~notions that define fairness using a similarity measure; (3)~notions that define fairness through causal modeling; and (4)~notions that define fairness based on utility. These are reviewed in the following. At this point, we emphasize that the fairness notions below have been primarily developed for predictive analytics. We highlight exceptions that explicitly consider prescriptive models and discuss implications for algorithmic fairness in business operations.

\subsubsection{Measuring fairness at group level.}

Fairness at group level, also referred to as statistical notions of fairness, quantifies statistical disparities across groups, where the groups are typically defined in terms of sensitive attributes, such as gender or race \citep[\eg,][]{Chouldechova.2017,Hardt.2016,Kleinberg.2017}. Defining fairness at an aggregate level based on a sensitive attribute is in part motivated by legislative frameworks, where such approaches are common \citep{Barocas.2016}. In this section, we provide a review of three commonly used metrics of group fairness: statistical parity, equalized odds, and predictive parity. 

Importantly, across different contexts, it is useful to appreciate that discrimination might arise against individuals at the intersection of multiple (sensitive) attributes, even when no group characterized by a single attribute experiences discrimination. Hence, the notion of a sensitive attribute in such contexts ought to be defined thoughtfully to reflect these realities. For instance, at General Motors, hiring practices were found to be biased against black women \citep{WashingtonPost.2015}. This discrimination was not visible; however, when inspecting the hiring of only women (mostly white office clerks) or of only black employees (mostly men hired as factory workers).

\textsc{Notation.}  Let $h$ denote a model. Based on input $X$, the model makes a prediction $\overline{Y} \in \{ 0, 1 \}$, while $Y$ is the actually observed outcome. We refer to $Y = 1$ as the positive outcome and to $Y = 0$ as the negative one. Without loss of generality, we assume a sensitive attribute $A$ that defines a disadvantaged group, $s_1$, and a group of individuals not belonging to the disadvantaged group, $s_2$. We write $Z_1 \indep Z_2$ if two random variables are independent.

\begin{figure}[ht]
\centering
\includegraphics[width=.8\linewidth,valign=t]{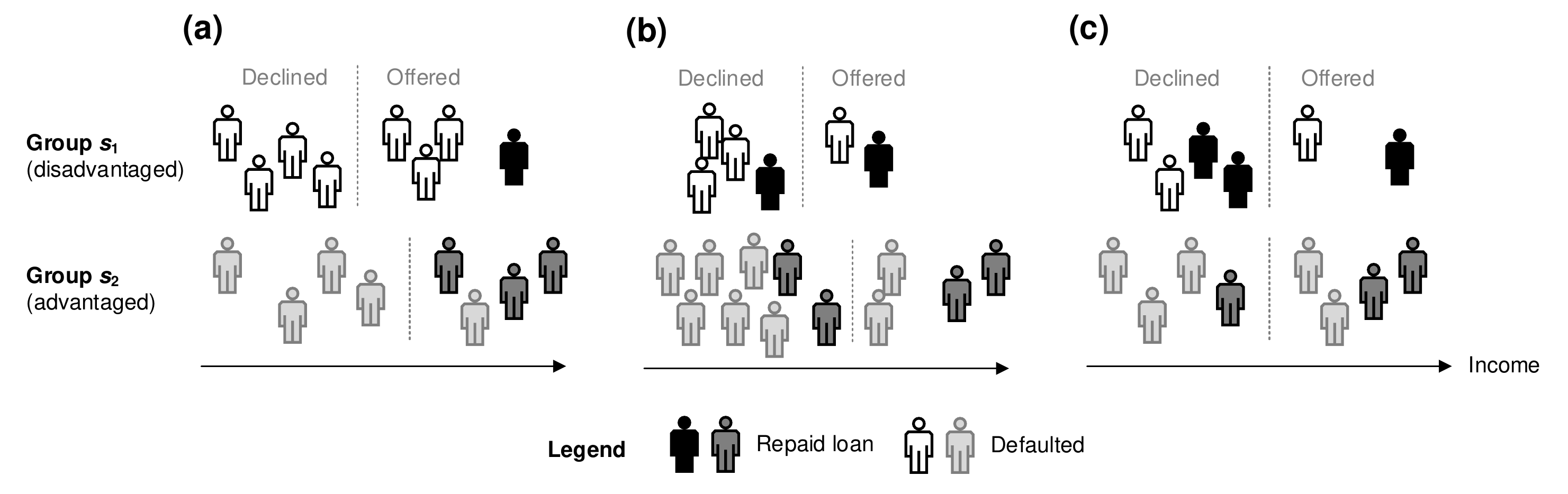}
\SingleSpacedXI\scriptsize
\begin{tabular}{p{16cm}}
\emph{Notes:} Example based on credit risk modeling. The gray dotted line represents the decision threshold. Individuals left of it will be declined a loan, while individuals on the right-hand side will be offered a loan. In~(a), the percentage of people offered a loan is equal (\ie, 50\,\%) for both groups. Thereby, actual defaults are not considered. In~(b), equalized odds ensures that positive outcomes are correctly predicted at an equal rate across groups, and that false positive outcomes are made at equal rates across both groups. Here, for both groups, 50\,\% of those who would repay (dark color) are correctly predicted to be creditworthy and are thus offered a loan. Moreover, 25\,\% of those who would default (light color) are incorrectly assumed as creditworthy and thus offered a loan. In~(c), predictive parity is ensured, as the precision is 50\,\% for both groups (\ie, 50\,\% of the loans are repaid).
\end{tabular}
\vspace{0.2cm}

\caption{Illustration of three fairness notions: (a)~statistical parity, (b)~equalized odds, and (c)~predictive parity.}
\label{fig:fairness_metric}
\vspace{-0.6cm}
\end{figure}

\emph{Statistical parity} is a common notion of fairness. Sometimes also called demographic parity, it entails that the positive outcomes are generated with equal probability across groups; see \Cref{fig:fairness_metric}(a). An example is a decision model for hiring decisions that should eventually offer jobs with equal probability to women and men. Thus, $P(\overline{Y} = 1 \,\mid\, s_1) = P(\overline{Y} = 1 \,\mid\, s_2)$ should hold true. Formally, it ensures $\overline{Y} \indep A$, that is, the prediction is independent of the sensitive attribute $A$. Notions similar to statistical parity are common in legal frameworks, in approaches such as the 80 percent rule \citep{Barocas.2016}, which considers a sufficiently large difference in selection rates across groups to be a presumption of adverse impact. Unlike the other notions below, statistical parity is based only on the predicted outcome, not considering the actually observed outcome (\ie, whether a hire succeeded at a job). This can be a key weakness in practice, as it ignores any possible correlation between $Y$ and $A$, which might not be desirable in cases of different base rates. 

\emph{Equalized odds} accounts for the fact that the classifier is subject to prediction errors and that the prediction errors might be biased against a disadvantaged group. This can happen when certain prediction errors (\ie, type-I and type-II errors) are over-proportionally frequent for the disadvantaged group; see \Cref{fig:fairness_metric}(b). For example, in a decision model predicting loan default, equalized odds would imply that the same ratio of correct offers/declines is made for each group. Formally, this notion corresponds to the metric that ensures the interdependence between the sensitive attribute and both type-I/II errors \citep{Hardt.2016}. Specifically, this is formalized by $\overline{Y} \indep A \,\mid Y$. In practice, equalized odds is particularly relevant for decision-making problems in which a positive prediction grants a specific benefit to individuals, yet where errors in granting this benefit should be equal across groups, as disparities are perceived as unfair or place a burden on individuals.

\emph{Predictive parity} stipulates that the precision rates are equivalent for all subgroups under consideration; see \Cref{fig:fairness_metric}(c). It thus ensures that the probability of a correct prediction is equal for all values of the sensitive attribute, regardless of whether an individual belongs to the disadvantaged group or not \citep{Chouldechova.2017}. As an example, let us consider credit risk modeling for men and women. Under predictive parity, whenever a model predicts a loan would be repaid and thus grants a loan, the rate of default should be the same across groups. Formally, in predictive parity, individuals who receive a predicted probability of $p$ should have a fraction $p$ of positive outcomes regardless of the group (\ie, $Y \indep A \,\mid\, \overline{Y}$). Owing to this, such a predictive model is also referred to as being calibrated across groups. BA applications where this is often perceived as desirable include risk assessments provided to human experts in domains such as supply chain and healthcare, since it implies that the estimated risk has a similar meaning across all subgroups.

\subsubsection{Measuring fairness via a similarity measure.}

An alternative is to define fairness at the individual level, seeking to ensure that individuals with similar properties also receive similar outcomes \citep{Dwork.2012}. This notion of fairness makes comparisons amongst all individuals in the data, regardless of a group membership. Because of this, an a~priori specification of sensitive attributes is not needed; instead, all available covariates (or a designated subset thereof) are used for comparisons. A potential application is automated hiring processes, where fairness at the individual level ensures that two job candidates with similar profiles should have the same chance for employment. 

Formally, let $d(x_i, x_j)$ define the distance between two individuals and let $D$ measure the distance of two distributions. Then, a decision model $h$ (\eg, a predictive or a prescriptive model) should yield similar outcomes for similar input, \ie, $D(h(x_i), h(x_j)) \leq d(x_i, x_j)$. For some BA tasks, using a notion of individual-level fairness could yield a flexible way to formalize non-disparate outcomes. This holds true even when it is challenging to define a set of sensitive attributes. However, individual-level fairness also has an inherent challenge: what is a task-relevant way of measuring similarity across individuals? This is challenging--and oftentimes even impossible--to operationalize in practice (\eg, comparisons across discrete covariates are non-trivial and highly domain-specific).

\subsubsection{Measuring fairness through causal modeling.}

A different approach towards identifying algorithmic bias approaches the problem through a causal lens. For this, a prerequisite is a causal model that is tailored to the BA task. Such a causal model can be obtained in two ways: either by specifying the underlying causal graph or by learning it from data. Once the causal graph is specified, the presence of discrimination can be formalized in terms of the pathways through which the sensitive covariate has an effect on the outcome \citep{Nabi.2018}. For instance, if women are less likely to be hired as construction workers by a company, physical strength could be considered an admissible pathway whereas having kids could be considered an inadmissible one. While this is a very intuitive and desirable formulation, there is an inherent challenge: the causal graph must be a~prior known; yet specifying a causal graph is often challenging, requires domain expertise, and, for many applications, might be subject to debate.  

Alternatively, counterfactual notions of fairness define a model as fair if the model's estimated outcomes would be the same if the individual belonged to a different demographic group \citep{Kusner.2017}. Importantly, this entails not just changing the value of the sensitive attribute, but changing other information that is causally dependent on that attribute. Machine learning approaches have been proposed to estimate the unobserved outcome for an individual that would hypothetically belong to the other group, \ie, the so-called counterfactual \citep{Kusner.2017}. By comparing the observed and the counterfactual outcome, the causal effect of a sensitive attribute on a BA outcome can then be estimated. In the absence of bias, there should be no causal effect of a sensitive attribute on the outcome. However, the validity of such counterfactuals has been brought into question, since attributes such as gender and race derive their social meaning from many of the features that such frameworks treat as their ``effects'' \citep{Hu.2020}.

\subsubsection{Measuring fairness based on utility.}

A different research stream has developed fairness notions that account for the utility to individuals who are subject to a decision. For this, a key prerequisite is a~priori knowledge of a utility function $U$, which describes the preference of an individual toward certain allocation of goods.

Using a given utility function, an allocation fulfills \emph{proportional fairness} if, compared to any other feasible allocation of utilities, the aggregate proportional change in the utilities is less than or equal to zero \citep{Bertsimas.2011}. Intuitively, an allocation is proportionally fair when any transfer of resources would lead to a relative benefit in the utility of one individual but reduce the utility of another individual by the same magnitude. Another notion is \emph{max-min fairness}, which seeks to maximize the minimum utility that any individual obtains \citep{Bertsimas.2011}. Formally, proportional fairness can be seen as the generalization of the Nash solution for a two-player problem, and max-min fairness is often grounded in Rawlsian justice and the Kalai-Smorodinsky solution in the two-player problem. Both have been used for studying the impact of fairness in resource allocation \citep{Bertsimas.2011,Rea.2021} including decision models that support, \eg, organ transplantation \citep{Bertsimas.2013}. Of note, \citet{Cohen.2019} extend fairness notions to pricing. A third notion of utility-based fairness terms allocations \emph{envy-free} if an individual receives an allocation that has the same (or a higher) utility as the allocation of any other individual \citep{Balcan.2019}. One can also adapt envy-freeness to predictive settings, where it warrants that predictions are then envy-free with high probability \citep{Balcan.2019}. 

However, the aforementioned fairness notions introduce a key challenge for OM research and practice: they require access to an individuals' utility function, which, in real-world applications, is typically difficult to infer from historical data or simply unknown.

\subsection{On Impossibility Theorems}

A key challenge in research and practice is that not all fairness notions can be fulfilled at the same time. In fact, it is mathematically impossible for these to be simultaneously satisfied. Under certain conditions such as different base rates, the notions of (i)~statistical parity, (ii)~equalized odds, and (iii)~predictive parity are mutually exclusive for a well-calibrated classifier \citep{Chouldechova.2017,Friedler.2016,Kleinberg.2017,Pleiss.2017}. Hence, seeking fairness with respect to one fairness metric will necessarily increase disparity with respect to the others \citep{Pleiss.2017}. Consequently, if more than one fairness notion is of importance, one could only attempt to find relaxations of fairness guarantees that are the mathematically feasible \citep{Kleinberg.2017,Pleiss.2017}. 

\subsection{Guidance on How to Choose Fairness Notions}

The need for manually choosing a suitable fairness notion can expose tensions in the goals and values of different stakeholders \citep{Coyle.2020}, as underpinned in the heated debated around the COMPAS system, a criminal risk assessment tool deployed in the U.\,S. Some argued that the COMPAS system is biased toward black defendants because it was more likely to \emph{incorrectly} flag black defendants than white defendants as future offenders \citep{Angwin.2016}. Mathematically, the tool had a higher false positive rate for black defendants, violating equalized odds. Meanwhile, the company that developed the algorithm argued that COMPAS satisfied predictive parity \citep{Chouldechova.2017}. That is, scores had the same calibration across groups, so for a given score threshold, the error rate was the same regardless of race. Importantly, whenever base rates are different (\eg, different recidivism rates across groups), equalized odds and predictive parity cannot be simultaneously satisfied \citep{Chouldechova.2017}.

For both researchers and practitioners, these impossibility results have important implications: a choice for a suitable metric must be made. To help inform this choice, we offer below a series of questions that provide guidance, as there is no one-size-fits-all solution (\Cref{tbl:choice_fairness_metric} lists \mbox{(dis-)}advantages of the different metrics). Furthermore, it is crucial to understand that no metric is an end in itself, and its use must be grounded on context-dependent goals and values that it aims to operationalize.
\begin{itemize}
\item \emph{What are the goods and burdens allocated by the possible decisions?} If the task entails the allocation of a burden (\eg, public sector operations such as denying bail), concerns over differential false positive rates may be well-justified. Meanwhile, in a task such as predicting someone's occupation to show them relevant job postings, the allocation of a good (\eg, showing a relevant job) may justify a focus on differential false negative rates \citep{DeArteaga.2019}. The metric in use should focus on the relevant disparities of goods/burdens in a given context.
\item \emph{What are the previous harms and injustices that the algorithm risks compounding?} In lending, for example, historical financial exclusion may be compounded by differential false positive rates in the prediction of loan default. Meanwhile, gender discrimination in certain occupations could be compounded if women in such an occupation are less likely to be recruited for hiring (higher false negatives) \citep{DeArteaga.2019}. Concerns over algorithmic unfairness are grounded on the risk that algorithms may perpetuate societal biases and compound historical discrimination \citep{Hellman.2018}.
\item \emph{How will the choice of metric affect other desiderata?} Algorithms do not operate in a vacuum, and choices grounded on fairness considerations may inadvertently affect other desiderata. For instance, if the algorithm is embedded as a decision support tool that provides probability estimates to a human expert, then calibration--and consequently predictive parity--may be desirable \citep{Chouldechova.2017}. Meanwhile, this may be a less pressing issue for algorithms whose output interpretation does not hinge on calibration (\eg, algorithms that ``flag'' a subset of cases while making no inference on the remaining instances), eliminating tensions with equalized odds. 
\end{itemize}

\begin{table}[ht]
\centering
\caption{Guidance for choosing fairness metrics. }
\label{tbl:choice_fairness_metric}
\SingleSpacedXI
\tiny
\def\arraystretch{1.3}
\begin{tabular}{p{1.5cm} p{2.5cm}p{2cm}p{9.5cm}}
\toprule 
\textbf{Fairness metric} & \textbf{Setting} & \textbf{Use cases} & \textbf{Advantages/disadvantages} \\ 
\midrule 
\multicolumn{4}{l}{\emph{Fairness at group level}} \\
Statistical parity & Predictions or decisions with given sensitive attribute & Hiring or other selections requiring demographic parity & \fplus~Ensures equal selection rates across groups; straightforward to audit and assess for regulators; does not require observed outcomes \newline
\fminus~Considers only the predicted outcome, not the actual outcome, hence does not differentiate between incorrect and correct predictions/decisions; requires a predefined sensitive attribute, which may be not known/defined \\
Equalized odds & Predictions or decisions with given sensitive attribute & Risk assessments (\eg, in supply chains and healthcare) & \fplus~Ensures that error rates for different types of errors (\eg, false positives) are similar across groups, thus differentiating between types of possible harms \newline
\fminus~Requires a predefined sensitive attribute, which may be not known/defined; if base rates are different, scores across groups will not be calibrated \\
Predictive parity & Predictions or decisions with given sensitive attribute & Risk assessments (\eg, in supply chains and healthcare) & \fplus Ensures calibration across groups, so meaning of a risk score can be interpreted similarly across groups, which is important when risk scores are used for subsequent decision-making \newline
\fminus Requires a predefined sensitive attribute, which may be not known/defined; it does not differentiate between different types of errors \\
\midrule
\multicolumn{4}{l}{\emph{Fairness via similarity measure}} \\
Individual fairness & Well-defined similarity measure & & \fplus~No need for choosing sensitive attributes; works for predictive and prescriptive models \linebreak
\fminus~Computationally challenging, as it allows only for approximate solutions; it is hard to define an appropriate similarity measure  \\
\midrule
\multicolumn{4}{l}{\emph{Fairness through causal modeling}} \\
Causal graphs & Causal graph is known & Behavioral research & \fplus~Powerful for expanding the theoretical understanding in behavioral sciences \linebreak 
\fminus~Causal graph typically unknown and thus infeasible \\
Counterfactual fairness & Sensitive attribute is known and estimated counterfactual carries meaning/significance & Auditing & \fplus~Counterfactual reasoning promotes traceability and accountability of decision logic \linebreak 
\fminus~Strong mathematical assumptions that, due to the absence of counterfactual outcomes, cannot be validated in practice
\linebreak
\fminus~Counterfactuals on sensitive attributes often lack significance (\eg, if the race/gender of someone changed, their whole lived experience would change, too)\\
\midrule
\multicolumn{4}{l}{\emph{Fairness based on utility}} \\
Proportional fairness & Decision problems with known utility function & Allocations (\eg, organ transplantation), scheduling, etc. & \fplus~Optimizes considering everyone's individual utility so that, in an allocation, no other solution could improve utility for an individual without reducing utility for others by a greater magnitude \linebreak 
\fminus~Utility function of individuals is rarely known \\
Max-min fairness & Decision problems with known utility function & Allocations (\eg, organ transplantation), queuing, etc. & \fplus~Optimizes considering individual utilities and focusing on those who are worse-off so that, in allocation, no other solution could improve utility for some without decreasing utility for those receiving lowest utility level \linebreak 
\fminus~Utility function of individuals is rarely known \\ 
\bottomrule
\end{tabular}
\end{table}

\section{Mitigating Algorithmic Bias in Business Analytics}
\label{sec:bias_mitigation}

In this section, we discuss mitigation of algorithmic bias in BA and conclude with guidance for practitioners on how to choose an effective bias mitigation strategy.

\subsection{Why It Is Ineffective to Omit Sensitive Attributes}

One might be inclined to think that a na{\"i}ve approach to achieving algorithmic fairness is by simply omitting all attributes deemed sensitive. For instance, to avoid racial bias, one may want to prevent the model from having access to a variable that contains individuals' race. However, the inclusion of sensitive attributes need not lead to unfairness \citep{Kleinberg.2018b}, and when it does, its exclusion does not necessarily address the problem \citep{Dwork.2012}. Access to sensitive attributes may improve fairness properties of an algorithm when the predictive relationship between variables and outcome varies across subgroups. Consider the task of algorithmic-assisted hiring based on covariates that capture education. As a result of socioeconomic disparities and barriers of access to resources, first-generation college graduates with the same potential to succeed at future jobs may graduate with lower grade average (\eg, GPA). By having access to a covariate that indicates whether someone is a first-generation college graduate, the algorithm may account for this bias by identifying that the predictive relationship between GPA and future job performance depends on this socioeconomic factor.

Access to sensitive attributes, however, may lead the algorithm to use these as a proxy for the target outcome in ways that compound previous disparities. For instance, if structural challenges have made it harder for first-generation students to graduate college or succeed at their first job, the algorithm may learn a negative association between first-generation graduates and job performance, leading it to predict that a first-generation graduate is less likely than others to succeed, even if everything else in their profiles is identical. In such cases, excluding the sensitive attribute is often not enough to address the problem. The reason is that many variables are correlated with sensitive attributes and thus act as proxies. For instance, students' zip code or school district may be used by the algorithm as a proxy for race. A prominent example where omitting sensitive attributes was ineffective is Amazon's free-same day delivery \citep{Bloomberg.2016}. Amazon declared that no data on race was used for deciding where to offer free delivery services. Nevertheless, many other covariates may act as potential proxies (\eg, the zip code), which resulted in a service that excluded predominantly black neighborhoods.

In sum, omitting sensitive attributes is neither necessary nor sufficient to prevent disparate outcomes in BA. Thus, algorithmic fairness approaches ought to be considered.

\subsection{Addressing Sources of  Algorithmic Bias during Data Collection}

One approach to mitigating bias in BA practice is to invest in \textquote{good data.} For this, several approaches can be considered. First, it is important to document how the data was recorded, including the data composition, collection process, and recommended uses \citep{Gebru.2018}. Such information can help evaluate whether it is appropriate to reuse data for a new task and identify potential sampling bias, differential subgroup validity, and other sources of bias that stem from data collection and representation (as described in \Cref{sec:sources_data}). Second, potential sources of bias in the data should be critically assessed by domain experts. This might imply that a BA task is eventually revised. For instance, one might need to quantify a business goal through a different proxy or to exclude data sources that are themselves biased. Third, it might be necessary to improve the data collection practices (or remedy errors in already collected data). For instance, if missing values in electronic health records appear with larger relative frequency for black patients, then it is useful to identify the cause and improve the underlying reporting practices \citep{Vokinger.2021}.  

In addition to careful design and documentation, it is also possible to make use of algorithmic fairness approaches to guide the collection of better data. For instance, selectively acquiring additional information in cases where the model lacks predictive power can guide the allocation of a limited budget in a way that improves performance for disadvantaged groups \citep{Cai.2020}. Active learning, which corresponds to computational techniques to identify informative data, can also be adapted to optimize for a fairness objective, in ways such as accounting for the different biases and costs of humans tasked with labeling the data \citep{Gao.2020}.

The above approaches to addressing sources of bias during data collection are important in BA practice, yet often are  insufficient on their own. In what follows, we discuss mathematical approaches that constrain outcomes of decision models to improve fairness. 

\subsection{Mathematical Approaches to Algorithmic Fairness}
\label{sec:mathematical_approaches_algorithmic_fairness}

Mathematical approaches to mitigating bias in BA can be broadly characterized as modifications to the model or the process of estimating a model, with the goal of finding a solution that has good performance with respect to a specified notion of algorithmic fairness. 

\textsc{Example.} Let us consider an example from human resource operations where a predictive model, $f : x \mapsto y$, such as a neural network, is used to select candidates with profile $x \in \mathbb{R}^N$ for hire, \ie, $y \in \{ 0, 1 \}$. A na{\"i}ve application of predictive analytics would optimize against historical data, that is, $\min_f \; \mathbb{E}[\ell(y, f(x))]$ where $\ell$ is a loss function (\eg, cross-entropy loss). To foster gender equality, assume that a company wants to ensure statistical parity in the form of the so-called 80\,\% rule \citep{Barocas.2016}, so that the selection rate for women and non-binary candidates is no less than 80 percent of the selection rate for men. Then, one can adapt the above minimization to further incorporate a fairness constraint, with $\epsilon = 0.8$, 
\begin{equation}
\big\lvert  P(y = 1 \,\mid\, x_{\text{gender}} = \text{\textquote{man}}) - P(y = 1 \,\mid\, x_{\text{gender}} = \text{\textquote{woman or non-binary}}) \big\rvert \leq \epsilon
\end{equation}

Directly optimizing for the constrained optimization problem (\ie, in-processing) is one way of ensuring the desired rule is met. Alternatively, one could apply pre-processing to the data to account for possible biases prior to solving the optimization problem, or apply a transformation to the solution of the unconstrained optimization (\ie, post-processing). In the following, we discuss approaches to algorithmic fairness for predictive and prescriptive analytics.

\subsubsection{Algorithmic fairness in predictive analytics.}
\label{section:algfairness_pred}

In predictive analytics, approaches to algorithmic fairness can be generally grouped into how the fairness constraint enters the model. This corresponds to approaches for pre-processing, in-processing, and post-processing as follows.

\emph{Pre-processing} changes the data before the actual modeling step, so that the fairness constraint is later ensured. This can be done by reweighting, where the frequency of observations from the minority group is artificially increased through resampling and thus the observations from both disadvantaged and advantaged group become more balanced \citep{Kamiran.2012}. Alternatively, it is possible to learn representations of the data that exclude information that one does not wish a model to learn from. For instance, one can aim to obfuscate any information that correlates with membership to a disadvantaged group \citep{Zemel.2013}. 

\emph{In-processing} incorporates the fairness goal directly in the modeling objective \citep{Kamishima.2012,Wang.2020,Zafar.2019}. This can be achieved by reformulating the model through an optimization constraint, which limits the space of acceptable utility-optimizing solutions to a set that meets certain fairness criteria (\eg, modifying the optimization problem in a support vector machine through this additional fairness constraint). Alternatively, one can add the fairness constraint as a regularization term. While a constrained optimization ensures that the constraint is strictly met, the latter has the advantage that it generally gives a loss function that is differentiable and that can thus be solved by gradient-based solvers. Naturally, one can reformulate a constrained optimization in terms of a regularized objective via the Lagrangian form (or similar to the Big-M method in optimization). A different approach is to solve both the original objective and the fairness constraint through a game-theoretic approach. Specifically, one can formulate it as a minimax game that can then be estimated via adversarial learning \citep{Zhang.2018}.

\emph{Post-processing} takes an estimated model and changes its output in a way that the fairness constraint is post~hoc ensured. For instance, one can calibrate a model post~hoc toward equalized odd by applying Platt scaling \citep{Pleiss.2017} or by formulating an optimization problem that considers the output of the model and transforms it according to both overall performance and fairness constraints \citep{Hardt.2016}.

\subsubsection{Algorithmic fairness in prescriptive analytics.}

The above approaches to bias mitigation focus exclusively on predictive analytics. By contrast, in prescriptive analytics, algorithmic fairness is addressed in a way that not only the predictions but the actual decisions adhere to the fairness constraint. A na{\"i}ve approach is to map a prediction onto a decision using a probability threshold \citep{CorbettDavies.2017}. The threshold itself can be subject to optimization according to properties of the decision process, such as the tolerance for false alarms or resource constraints that determine the volume of cases that can receive a certain treatment or decision. 

Other approaches to algorithmic fairness consider specific model classes that are frequently used in OM. For instance, fairness constraints have been incorporated into a classic bandit \citep{Joseph.2016}. This presents a theoretical framework that can be used by OM researchers and practitioners to tailor fair bandits to domain applications (such as bandits in revenue management). Similarly, fairness constraints have also been considered in Markov decision processes \citep{DAmour.2020b}. However, the aforementioned works cover classic, unconstrained versions of the models, while more work is needed to adopt these to OM problems (\eg, one could adapt a bandit for an OM setting by incorporating resource or budget constraints). Finally, several works have studied the impact of fairness constraints specifically in an OM context. For example, some works incorporate fairness in resource allocation \citep{Bertsimas.2011,Bertsimas.2013,Rea.2021}, pricing \citep{Cohen.2019}, appointment scheduling \citep{Qi.2017}, and in supply chain coordination \citep{Katok.2014}. For future work, we foresee various opportunities to integrate fairness in models that fall under the scope of prescriptive analytics where optimization and machine learning are combined.

\subsection{Long-Term Effects of Algorithmic Fairness}
\label{sec:longterm}

Many approaches for algorithmic fairness are tailored to static settings in which long-term effects and system-level dynamics are not considered. In contrast to that, dynamic settings might be characterized by a utility that changes over time (\eg, due to positive or negative externalities) or due to additional long-term costs that need to be addressed. Whenever decisions at one time step affect data in the next time step, both utility and fairness may exhibit complex dynamics. 

There are recent efforts to model the long-term impact of algorithmic fairness. One stream of the literature adapts the loss function so that it accounts for feedback loops \citep{Hashimoto.2018}. Specifically, the authors employ distributionally robust optimization to minimize the worst-case risk over all distributions close to the empirical distribution. By doing this repeatedly in each time step, the approach controls the risk of bias for the disadvantaged group; however, it assumes that the data is sampled from the same distribution. Another stream in the literature models the effect of bias through a (sequential) decision-making framework, so that the long-term reward can be captured. For example, \citet{DAmour.2020b} develop a decision model in which the actions--with and without algorithmic fairness--are evaluated in a dynamic decision setting. The authors develop a Markov decision process that explicitly accounts for feedback loops in sequential decision-making. In some BA tasks, it might also be necessary to account for delayed impact \citep{Liu.2018}, as there may be a time lag until when disparities become quantifiable. For example, the financial impact of discrimination during college admission can only be measured several years after graduation.

\subsection{Guidance on How to Mitigate Bias in BA Practice}

\begin{table}[tbp]
\centering
\caption{Guidance for choosing a bias mitigation strategy. }
\label{tbl:guidance_mitigation}
\SingleSpacedXI
\scriptsize
\def\arraystretch{1.3}
\begin{tabular}{p{3cm}p{13cm}}
\toprule 
\textbf{Source of bias}&  \textbf{Mitigation recommendations} \\ 
\midrule 
\multicolumn{2}{l}{\emph{Data collection and representation}} \\
Sampling bias (\S\ref{sec:sampling_bias}) & \checkmark~Revised strategies for data collection to address sampling bias may be most effective \newline
\checkmark~Training models with fairness constraints or post-processing the outcome may be helpful if sampled data has enough signal for inference for under-sampled group(s)\newline
\faWarning~Additional data collection should be done ethically and with informed consent\\
Differential subgroup\newline validity (\S\ref{sec:differential_subgroup_validity}) & \checkmark~Modifications to the algorithm may be most effective to capture cross-group heterogeneity \newline
\faWarning~Using different predictive patterns for different groups may constitute disparate treatment and be at odds with some regulation \\
Biased observed \newline outcomes (\S\ref{sec:biased_observed_outcomes}) & \checkmark~Revisiting problem formulation may be most effective, especially by acknowledging gap between desired outcome and observed outcome \newline
\checkmark~Data collection may be effective if another (less biased) outcome can be collected \newline
\faWarning~Training constraints and post-processing approaches may be ineffective and yield misleading results because most of these rely on the assumption that observed outcome is the true outcome\\  
\midrule
\multicolumn{2}{l}{\emph{Model estimation}} \\
Optimization objectives and evaluation bias (\S\ref{sec:sources_objective}) &  \checkmark~Controlling bias through modifications to the algorithm likely most effective \newline
\faWarning~The ability to appropriately diagnose this may be complicated by an over-reliance on chosen performance metrics\\
Bias ingrained in the assumptions of the \newline objective (\S\ref{sec:bias_assumptions_objective}) & \checkmark~Revisiting the problem formulation is likely necessary \newline
\faWarning~Flaws in problem formulation may imply a `non-solution' and require a substantial restructuring, which may include deciding that a technology should not be built  \\
\midrule
\multicolumn{2}{l}{\emph{Deployment}} \\
Bias in BA adoption (\S\ref{sec:deployment}) & \checkmark~Modifications to the algorithm may be effective if they improve human-AI complementarity \newline
\checkmark~Implement post-deployment monitoring to routinely probe for bias and potential feedback loops \newline
\faWarning~Appropiately identifying and mitigating bias requires assessing the human-AI team; considering the algorithm in isolation will not suffice \\
\bottomrule
\end{tabular}
\end{table}

Managers have four fundamental ways to address bias in BA practice: (1)~control the source of bias at the problem formulation stage, (2)~control the source of bias at the data level, (3)~control the bias through modifications to the algorithms, and (4)~control the bias by post-processing the output. Which mitigation strategies are effective will depend heavily on the source of bias, and we thus provide guidance for practitioners in \Cref{tbl:guidance_mitigation}. Therein, we build upon the sources of bias from in~\Cref{sec:bias_sources} and, to address them, suggest mitigation strategies.
\begin{itemize}
\item \emph{There is no silver bullet.} Instead, a combination of different strategies is oftentimes needed. Consider the example of clinical trials involving insufficient racial diversity \citep[\eg,][]{Warren.2020}. As a way forward, it is ethically indicated to conduct clinical trials that are representative of multiple races. Nevertheless, it is unfeasible to replace all existing clinical trials, and it is thus important to adjust for the selection bias through computational approaches when using the existing data from previous trials. As a rule of thumb, companies should continuously check new data collection efforts for risk of bias, revisit problem formulations, and critically inspect baked-in assumptions, while an appropriate diagnosis of sources of bias can enable mitigation via mathematical approaches to algorithmic fairness when appropriate, allowing managers to produce data-driven insights from existing data sources (\Cref{sec:mathematical_approaches_algorithmic_fairness}). 
\item \emph{Mitigation strategies need to consider business context.} Here, appropriate domain understanding and engagement of affected stakeholders is crucial, as this is necessary to enable the diagnosis -- and mitigation -- of bias. Consider the example of FinTechs that leverage novel data sources for credit scoring. Here, knowledge of the business model and process is necessary (but may not be sufficient) to assess which data sources are at risk of bias. Complementing domain knowledge with community engagement via participatory design strategies can help anticipate modes of failure, reshape problem formulation, and guide the implementation of effective bias mitigation strategies. 
\item \emph{Auditing.} The different strategies for bias mitigation vary in the extent to which they facilitate auditing. Consider the above selection bias example, which a company may want to mitigate via a new data collection practice. Here, regulators can assess whether the bias mitigation is successful by inspecting whether the data is now representative. In contrast to that, auditing mathematical approaches to algorithmic fairness (\Cref{sec:mathematical_approaches_algorithmic_fairness}) can bring challenges by requiring specialized mathematical knowledge and novel auditing strategies when the goal is to audit the \emph{process}. A growing number of software tools aim to support practitioners in this task (we provide an overview in the e-companion).  
\end{itemize}

\section{Fairness-Utility Relationship}
\label{sec:fairness-utility}

Conversations about fairness in OM and business applications are often accompanied by a discussion on trade-offs and the price of fairness \citep{Bertsimas.2011,Bertsimas.2012, Bertsimas.2013,Cohen.2019,Katok.2014,Rea.2021}. Typically, it is assumed that fairness constraints will restrict the search space of a decision model and will thus come at a cost to utility, where utility is understood as referring to traditional business goals (\eg, profit, cost). For instance, in the context of airline operations and flight rescheduling due to inclement weather, the cost-minimizing solution incurs higher costs for some airlines than for others. When one ensures that no airline is disproportionately affected, the total cost from delays becomes larger \citep{Bertsimas.2011}. However, in some settings, improving fairness may also lead to an improvement in utility. Consider a bank that systematically underestimates women's financial standing and likelihood of repaying a loan. This misestimation is not only harmful to women, but it is also harmful to the bank itself, which is missing out on potential profits associated to these loans. 

The assumption of a trade-off between utility and fairness is intuitive from an optimization perspective \citep{Bertsimas.2012}; but claims of an inherent trade-off implicitly make the following four assumptions: (1)~the utility function is strictly convex; (2)~there is no sampling bias; (3)~the optimization objective corresponds to the business utility \emph{and} there is no mismeasurement error; and (4)~there are no feedback loops between the effects of algorithmic deployment and future values of the utility function. Importantly, in many cases, the sources of bias are precisely related to a violation of some of these assumptions.

Understanding the utility-fairness relationship is fundamental to a successful characterization and incorporation of fairness considerations in business practices. In the first part of this section, we discuss cases in which fairness is not at odds with utility, a phenomenon that is not rare but is frequently underappreciated. In the second part, we discuss cases where there is or may be a trade-off, and the business implications this has. 

\subsection{Absence of a Fairness-Utility Trade-Off}

Characteristics of the utility function, the model's ability to correctly estimate it, and the dynamics of deployment may all lead to an absence of a fairness-utility trade-off. 

\begin{figure}[h!]
\centering
\vspace{-0.3cm}
\includegraphics[width=0.5\linewidth]{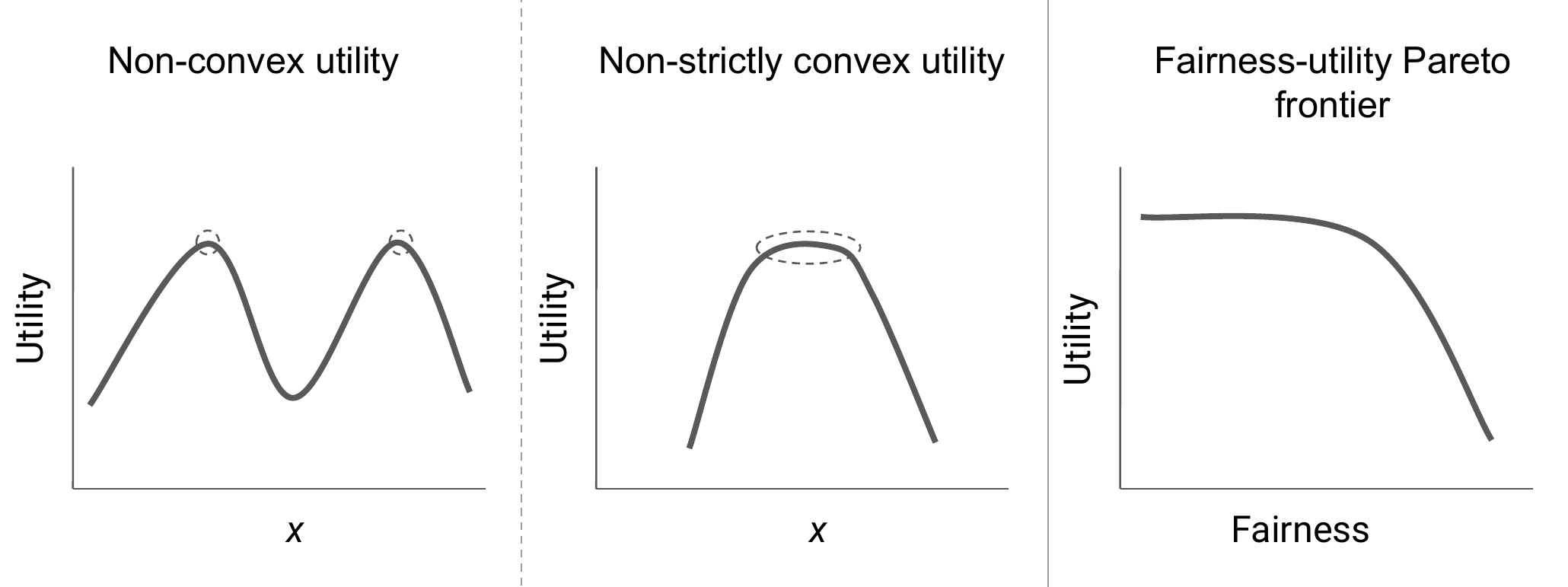}
\vspace{-0.3cm}
\caption{Illustrations of non-unique solutions (left and center), and corresponding Pareto frontier (right), which yields multiple solutions that are equivalent from a utility perspective.}
\vspace{-0.5cm}
\label{fig:nonunique}
\end{figure}

\paragraph{Non-unique solutions.} The solution to an optimization objective is often non-unique. In the context of predictive modeling, this phenomenon has been studied as underspecification \citep{DAmour.2020} or predictive multiplicity \citep{Marx.2020}. In such settings, solutions that are equivalent from a utility-maximizing perspective can be very different from a fairness perspective \citep{Chouldechova.2017b, Marx.2020}. As illustrated in \Cref{fig:nonunique}, non-unique solutions may arise when the optimization objective is non-convex, or it is convex but not strictly convex. In such settings, the utility-fairness Pareto frontier may have multiple solutions that yield the same utility, but are different from a fairness perspective. For instance, in the context of automated and targeted recruiting, researchers have found gender bias in the task of predicting individuals' occupations from their online presence. The same research also shows that removing explicit gender indicators from biographies can mitigate (although not remove) the observed gender gap, while maintaining the overall prediction accuracy \citep{DeArteaga.2019}. The existence of non-unique utility-maximizing solutions has two main implications: (1)~in such settings, significant gains can be made from a fairness perspective without incurring a utility loss, and (2)~such gains are likely to not be realized if the optimization objective does not explicitly encode fairness considerations. 

\paragraph{Bias in utility estimation.} As discussed in \Cref{sec:bias_sources}, the mechanism through which an algorithm learns a biased solution is often a misestimation of the utility function. Flawed proxies can result in unfair decisions, but they may also lead to sub-optimal solutions from a utility-maximizing perspective \citep{Bastani.2021c}. For instance, consider the predictive model meant to estimate health needs to target patients for high-risk care management \citep{Obermeyer.2019}. A utility-maximizing solution would be one that correctly predicts who are the high-risk patients. Thus, a bias mitigation intervention that corrects for the misestimation of needs for black patients has the potential of improving both fairness properties and utility maximization.

Finally, accurate utility estimation also relies on properties of the data distribution. Consider the case of models that are fitted with convenient, yet non-representative samples. During estimation, a utility-maximizing solution may be one that sacrifices performance for an under-sampled population in favor of an over-sampled one. Yet, this may not maximize utility in deployment, when the shift in data distribution could yield costly errors for the initially under-sampled population.

\paragraph{Unmeasured or dynamic costs.} Even within the narrowly defined task of utility maximization, there are often unmeasured costs and benefits that may impact the overall utility, such as employee satisfaction. Some of this unmeasured component of the utility may be directly linked to fairness properties of an optimization problem. For example, in the context of bias in algorithmic hiring \citep{Raghavan.2020}, unmeasured costs of that bias include the potential cost of having less diverse teams \citep{Page.2017}. Moreover, models developed considering a static setup may fail to capture dynamic or long-term costs of certain policies. If individuals invest rationally, unfairness in an algorithmic output will impact incentives, and may affect (rational) investment by affected stakeholders \citep{Liu.2020}. Similarly, costs to reputation and spillover effects on perceptions of fairness may affect profits \citep{Katok.2014}.

\subsection{Presence of a Fairness-Utility Trade-Off}

A fairness-utility trade-off may arise for a number of reasons. In this context, it is useful to differentiate between \emph{accuracy-affecting injustices} and \emph{nonaccuracy-affecting injustices} \citep{Hellman.2020b}. Below, we define each type of injustice, discuss why the trade-off may arise in each case, and what the implications for business operations are. 

\paragraph{Fairness-utility trade-off and accuracy-affecting injustices} Accuracy-affecting injustices refer to cases in which either the data or the output of a model \emph{inaccurately} estimate a fact about people \citep{Hellman.2020b}. Examples include underestimation of black patients' health needs \citep{Obermeyer.2019}, or of women's ability to repay a loan. In principle, these are cases in which fairness and utility should not be at odds; correcting the misestimation has the potential to increase the utility. In practice, that is not always the case. A fairness-utility trade-off arises when a solution that optimizes overall performance has poor performance for a subpopulation, \emph{and} alternative solutions necessarily reduce the overall performance. This may happen for a variety of reasons. A constraint in complexity of the model may mean that the solution that optimizes for overall performance underperforms for a subpopulation (\Cref{sec:sources_objective}), and, under the same complexity constraints, improving performance for that subpopulation could harm performance for the majority population. While alleviating the complexity constraint may seem like the obvious solution, this may involve other trade-offs, such as the risk of overfitting. A utility-fairness trade-off may also arise if the source of the bias is differential subgroup validity (see \Cref{sec:sources_data}), and the available data does not contain information that is predictive for a subpopulation. In this context, the only way of enforcing a fairness metric without collecting new data may be to reduce performance for the majority group. 

Notably, fairness-utility trade-offs in the context of accuracy-affecting injustices often stem from the constraint of relying on the same (flawed) data to estimate a model. In the context of business operations, this can often be circumvented by making a different choice of data or engaging in new data collection efforts. This can naturally help address unfairness resulting from sampling bias if the new data is more representative \citep{Raji.2019}. Choosing a different proxy outcome may also mitigate bias, because seemingly reasonable choices of proxy labels that yield similar accuracies can have markedly different implications in terms of bias, as has been observed in practice \citep{Obermeyer.2019}. Collecting more data may also help mitigate differential subgroup validity \citep{Cai.2020}. As an example, consider the task of deciding whether to grant applicants a loan. If a disadvantaged group is more likely to lack traditional lending histories, an algorithm that relies on this information may have lower accuracy for this group. This disparity in performance may be potentially reduced by allocating a limited budget to collect additional information for selected individuals (\eg, a time-consuming screening of non-traditional financial histories). While acquiring this information for everyone may be unfeasible, acquiring additional variables for selected individuals could improve the predictive accuracy for this subgroup, simultaneously improving the bank's ability to estimate utility and fairness of the decisions \citep{Cai.2020}.

\paragraph{Fairness-utility trade-off and nonaccuracy-affecting injustices.} Injustices may also arise when data and models \emph{accurately} estimate a fact about people, but these traits themselves result from injustice \citep{Hellman.2020b}. As an example, if racial discrimination is in part responsible for a racial pay gap, income would be a trait that results from that injustice. Even if not necessarily illegal, if racial disparity in allocations can be explained by this covariate, using it to inform decisions may still correspond to a form of indirect discrimination because it is informed by and compounds a previous injustice \citep{Hellman.2018}. 

The methodologies introduced in \Cref{sec:bias_mitigation} can help mitigate these type of injustices. For instance, enforcing demographic parity can ensure that a resource is equally distributed across groups. Examples of these type of algorithmic injustices in the context of business operations include the Staples online pricing case \citep{WSJ.2012} and the Amazon same-day delivery case \citep{Bloomberg.2016}. In the case of personalized online pricing, for instance, disparities do not cease to be problematic if the willingness to pay is accurately estimated. Similarly, the absence of same-day delivery in predominantly black neighborhoods was seen as a problem even if the company was grounding its decision based on (accurate) estimation of the volume of customers by zip code.

\section{Future Directions for Business Analytics Research}
\label{sec:future_directions}

More work is needed to detect and eventually mitigate bias in BA and thus shape an impactful research agenda around algorithmic fairness. To provide guidance in this task, we outline key research opportunities.

\subsubsection*{Quantifying the prevalence of algorithmic bias.}

As laid out above, algorithmic bias is widespread in BA practice, and it can arise across a variety of contexts and for different reasons. Media reports, government audits, and academic studies have provided empirical evidence showing instances of this phenomenon, but more research is needed to identify and document which industries are more at risk of introducing or perpetuating unfairness through the deployment of BA. Such inquiry can benefit from considering: Which operations and specific business decisions are at risk of algorithmic bias? Which stakeholders (\eg, customers, suppliers, individuals) have been historically disadvantaged and are therefore at higher risk of being impacted by algorithmic disparities? In light of this, we foresee great value for practitioners by estimating the prevalence of algorithmic biases across sectors, domains, and applications. This will eventually help practitioners in prioritizing applications that are especially prone to bias. Here, we hope that our discussion on the sources of algorithmic bias in \Cref{sec:bias_sources} provides a valuable starting point.

Eventually, such empirical evidence could raise awareness of the risk of algorithmic discrimination in the BA/OM community and further guide BA practitioners when auditing sources of bias to ensure a fair treatment of customers, suppliers, and other business stakeholders. This research would also enable the development of domain-specific guidelines for algorithmic audits in business operations, facilitating the robust and responsible deployment of BA technologies. 

The growing adoption of big data among business operations \citep{Choi.2018,Cohen.2018,Feng.2018} is likely to reinforce the risk of algorithmic bias. One reason is that algorithms for big data must be sufficiently scalable and will thus operate through heuristics, while algorithmic audits also become more computationally challenging. Companies may be especially prone to algorithmic bias when integrating data from social media in their business decision-making. Such data have many promising applications for businesses and thus offer large operational value \citep{Cui.2018,Lau.2018}. However, social media data are user-generated and thus reflect societal biases, which may affect downstream business decisions that, in turn, affect customers and suppliers. Moreover, possible biases and harms are not always easy to anticipate, as evidenced by Twitter's choice to create an ``algorithmic bias bounty challenge''\footnote{\SingleSpacedXI\scriptsize Twitter Blog (2021): \emph{Introducing Twitter’s first algorithmic bias bounty challenge}. URL: \url{https://blog.twitter.com/engineering/en_us/topics/insights/2021/algorithmic-bias-bounty-challenge}, last accessed October 18, 2021.} as a way to encourage users to uncover and report biases in their system. In the realm of big data and social media, an interesting path for future research is to explore how the operational value of social media changes when algorithmic fairness constraints are introduced.   

\subsubsection*{Measuring gains and losses with respect to status~quo.} 

Frequently, BA is meant to assist or automate decisions in operations previously performed by humans. The efficiency of algorithms is thus measured with respect to the previous status~quo. A similar approach is pertinent when considering the risk of bias \citep[\eg,][]{Kleinberg.2018c,Lu.2019,Yu.2020}. Thus, it is crucial to determine how bias of a model compares to previous bias embedded in the decision process. The nature of BA--learning from historical data--means that in many cases previous bias may be replicated and amplified. However, in other cases, algorithms may exhibit less bias than humans \citep{Ganju.2020}.

Crucially, how to measure and compare the bias across decision frameworks is a non-trivial question. The inconsistency present in human decisions, while often framed as a disadvantage, may also ameliorate the consequences of bias, while the  deterministic nature of algorithmic decisions could systematically exclude and harm people in ways that are mitigated by variance in human decisions. Thus, properly identifying the dimensions and measures along which to compare humans and algorithmic tools is an important question. In a similar vein, more research is needed that quantifies the efforts necessary for implementing algorithmic fairness in BA practices and thus to quantify the operational value. Such analysis must consider various different operational dimensions, including, \eg, costs from training personnel in algorithmic fairness, etc.

\subsubsection*{Mitigating algorithmic bias.} 

To mitigate algorithmic bias in BA, further research will be necessary along several directions. First, many of the above methods for algorithmic fairness have been developed for predictive analytics and not for prescriptive models. This suggests a substantial gap with exciting opportunities for research and practice to expand algorithmic fairness specifically in OM applications. In particular, there is no general purpose framework to developing such decision models. Hence, decision models with mathematical fairness constraints must be tailored to different types of optimization problems and formalized accordingly (\eg, matching markets, bandits, Markov decision processes). Future work can propose algorithmic fairness approaches for specific OM problems in areas not yet explored, including in impactful areas such as scheduling and routing. In addition, more theoretical analyses are needed, especially in the context of OM decision models (\eg, by deriving generalization bounds). Such theoretical findings will contribute to a better understanding of algorithmic fairness by informing decision-makers about what are potential (worst-case) implications.

In the big data era \citep[\eg,][]{Choi.2018,Cohen.2018,Feng.2018}, bias mitigation needs are dynamic and can evolve rapidly. As organizations increasingly rely on high dimensional and unstructured data and operate in multinational contexts with varied social backgrounds, it is important to develop methodologies that help practitioners discover what biases may be contained in a dataset, rather than having to probe for biases hypothesized a priori. Relatedly, one same product may have different fairness needs for different contexts in which it will operate, posing open questions for model training and deployment. Additionally, many data sources in the big data era of BA/OM are user-generated \citep[\eg, social media;][]{Cui.2018,Lau.2018} or collected via the Internet-of-Things (IoT), which necessitates a joint consideration of privacy and fairness. The increased use of varied sources of data has also given rise to new data sharing practices and machine learning paradigms, such as federated learning, opening new challenges and opportunities for bias mitigation. Mitigation strategies will further need to ensure accountability (\eg, for regulators) \citep{Feuerriegel.2022} despite involving high-dimensional and unstructured data hosted across locations, which poses computational and methodological challenges for auditing practices and policies.

Efforts in BA practice towards reducing algorithmic bias complement a larger paradigm shift in OM, calling for the importance of moving beyond profit maximization. \citet{Cachon.2020} state \textquote{We are most comfortable in the paradigm of optimization. But what should be optimized? Traditionally, we focused on costs and profits. But the objective functions of the future are more nuanced. Ethics, equity and well-being are of importance to many.} Thus, for the field of OM, it is important broaden the lens of what is and should be included in the optimization function as it will offer significant room for considering the role of business operations in mitigating--rather than compounding--historical injustices.

\subsubsection*{Evaluating dynamic effects and incentives.}

Most works on algorithmic fairness consider static settings; however, business environments are highly dynamic. Hence, care is needed for sequential decision-making or when the underlying population is subject to changes, thereby giving rise to potential feedback loops. To perform such evaluations, \citet{Kallus.2021b} propose a model for personalized pricing in which the long-run dynamics of fairness and utility are compared against a myopic revenue maximization. Similar analyses where the dynamic effects of algorithmic fairness are modeled could also be of interest for other BA tasks. 

When considering dynamic effects of interventions, an important success factor of algorithmic fairness is compliance \citep{Dai.2021}. If some firms adopt a non-compulsory, fairness-conscious policy and others do not, there may be complex interactive effects that affect the overall market. Similarly, customers might also adapt their behavior in response to how BA arrives at decisions affecting them \citep{Shimao.2019}. In this regard, a better understanding of equilibrium conditions in the context of multiple firms or players is important. Regulation may also have unanticipated effects on both utility and fairness if the incentives created by a policy are not properly understood \citep[cf.][]{Lambrecht.2019,Fu.2021}. Moreover, it is necessary to develop decision models that are robust to strategic manipulation. For instance, future research could leverage OM research on Stackelberg games for this purpose.

\subsubsection*{Designing BA for bias mitigation and human-algorithm complementarity.} 

As it has been emphasized, decision-makers may be biased too. This is increasingly relevant for research and practice, as the use of BA in some domains is shifting from automation to human-algorithm collaboration \citep{Bastani.2021b,Wang.2020,Senoner.2021}, including specifically to benefit fairness goals \citep{Wang.2020}. Further research on decision-making through human-AI collaborations can help achieve powerful complementarities, whereby both the limitations of human biases and of mathematical formalizations of fairness used towards fair BA can be alleviated, and where the relative strengths of humans and of algorithmic decision making can be leveraged to yield superior business operations. Ultimately, algorithmic fairness may provide a tool to mitigate human bias, but this necessitates empirical research to better understand how humans uptake algorithmic recommendations and methodological research that integrates these findings into better algorithmic design.

\subsubsection*{Quantifying fairness-utility relationship and measuring externalities.}

The above discussions around a potential trade-off are mostly theoretical, while actual evidence from the field that quantifies the costs and benefits of algorithmic fairness is absent. To fill this gap, OM could leverage its strength in performing experiments with companies with the goal of measuring the impact of algorithmic fairness in the field. Recent work showing that fairness-accuracy trade-offs can be negligible \citep{Rodolfa.2021} highlights the importance of empirical evidence to guide our understanding of the cost/benefit relationship of fairness-oriented efforts. This could not only quantify the financial implications but also allow researchers to study positive externalities such as customer perception, employee satisfaction, firm reputation, or competitive advantages, and measure operational costs of algorithmic fairness efforts. Moreover, the fairness-utility relationship has been primarily studied in static settings, yet fairness is likely to have a long-term impact on utility. Hence, a rewarding path for future research is to understand the long-term impact across organizational, economic, and societal dimensions in comparison to the status~quo.

\section{Conclusion}
\label{sec:managerial_implications}

The recent dramatic departure of business leadership from the view of business' principal responsibility as producing profits for shareholders, to the vision of business as having responsibility to all stakeholders, has lead to significant strides in the awareness and actions taken by business leaders towards fair business practices. Simultaneously, the fast integration of BA across industries has introduced an explosive risk of \textquote{bias at scale}: data-driven BA produces consistent business decisions, while automation of decisions and operations brings unprecedented scalability to the adverse impact of any bias in these actions. 

To date, meaningful efforts by firms towards algorithmic fairness in BA practices have been slow and scarce \citep{AINowInstitute.2019}, led primarily by the \textquote{big tech} companies, as exemplified by LinkedIn's deployment of algorithmic fairness as part of their ranking algorithm \citep{Geyik.2019}. Small- to medium-sized enterprises often lack the resources to make strides in their business operations. Indeed, some works aim at understanding current barriers to a widespread deployment of algorithmic fairness in organizations. In particular, interviews with analytics practitioners \citep{Holstein.2019} suggest that, perhaps not unexpectedly, more support is needed to guide practitioners in auditing models. 

Algorithmic bias can arise across a wide array of OM business contexts, plaguing both predictive modeling and prescriptive analytics, such as models used towards selection, scheduling, or resource allocation. Generally, discrimination is most likely to arise upstream for suppliers, and downstream for customers and users. Algorithmic fairness thus offers a wealth of opportunities for profound impact in practice, including possible impact on a firm's competitiveness and reputation for first-movers in particular. 

Moreover, an ongoing discussion in the OM field calls for an increased focus on objectives beyond profit-maximization. Algorithmic fairness is also relevant to a growing community of OM researchers and practitioners seeking to align OM with the United Nations' Sustainable Development Goals \citep{SDG.2015}, which includes the promotion of equality. Specifically, it calls for promoting equality and for the social, economic and political inclusion of all, irrespective of age, sex, disability, race, ethnicity, origin, religion, or economic status. Importantly, equality is not seen only as a basic human right, but also as a multiplier across all areas of sustainable development.

We expect that managers and researchers will engage significantly over the next decade in advancing the state-of-the-art towards objective functions in BA that are more nuanced and which specifically account for fairness. Bias mitigation calls for OM researchers and practitioners to develop models that are \textquote{fair by design.} Mitigating bias in human decisions has proven to be a formidable challenge. Meanwhile, algorithmic fairness provides ways for incorporating fairness goals into optimization, thereby yielding data-driven decision models that may better approximate fairness goals, especially over human decisions. However, it is important that OM researchers and practitioners remain aware of the limitations of mathematical formalizations to perfectly and comprehensively capture notions of fairness and remain vigilant of other injustices that may be invisibilized by a purely mathematical treatment of fairness.

Finally, we argue that OM's \emph{system and process} perspective on business value-creation positions it uniquely as a key discipline for advancing fair BA. In this paper, we describe how biased or fair BA can be driven by every element along the business-analytics process: from problem formulation, through data collection and algorithm development, to deployment and evaluation. The nature of the phenomenon calls for an overarching system and process perspective to identify, address, and critically evaluate solutions for challenges involving interactions of multiple elements in the BA process. While algorithmic fairness has received significant attention in other fields of analytics, including primarily in computer science, we hope that this paper makes the case for how research in BA and business operations is both necessary and rewarding. We also hope that this paper will help spark interest in fair BA research, facilitate business researchers introduction into the fascinating, non-trivial, and impactful challenges this area of inquiry presents, and inspire future OM research to follow the opportunities we outline and to make a positive impact in the world.

\section*{Acknowledgments}

{\SingleSpacedXI\footnotesize
Stefan Feuerriegel acknowledges funding from the Swiss National Science Foundation (197485). Maria De-Arteaga acknowledges funding from Good Systems (http://goodsystems.utexas.edu/), a UT Austin Grand Challenge to develop responsible AI technologies.  
}


\SingleSpacedXI
\renewcommand{\textquotedbl}{"}
\bibliographystyle{informs2014} 
\bibliography{literature}

\end{document}